\documentclass{article}

\usepackage[final, workshop]{neurips_2026}

\usepackage[utf8]{inputenc} 
\usepackage[T1]{fontenc}    
\usepackage{hyperref}       
\usepackage{url}            
\usepackage{booktabs}       
\usepackage{tabularx}       
\usepackage{amsfonts}       
\usepackage{nicefrac}       
\usepackage{microtype}      
\usepackage{xcolor}         
\usepackage{todonotes}     
\usepackage{amsmath}        
\usepackage{graphicx}      

\newcolumntype{Y}[1]{%
  >{\hsize=#1\hsize\raggedright\arraybackslash}X%
}

\usepackage{changes} 
\definechangesauthor[name={Kathrin}, color=red]{KK}
\definechangesauthor[name={Christian}, color=green]{C}
\definechangesauthor[name={Jenna}, color=orange]{J}
\definechangesauthor[name={Khoa}, color=blue]{KT}

\newcommand{\spec}{\ensuremath{\checkmark}}   
\newcommand{\est}{\ensuremath{\circ}}         
\newcommand{\unk}{\ensuremath{?}}             

\title{Implicit Assumptions About Tasks in Continual Learning}

\author{%
  Kathrin Korte \\
  IT University of Copenhagen\\
  Copenhagen, Denmark \\
  \texttt{kort@itu.dk} \\
  \And
  Christian Metzner \\
  Institute of Neuroinformatics \\ 
  University of Zurich and ETH Zurich\\
  \texttt{cmetzn@ini.ethz.ch}
  \AND
  Akshay Bedhotiya\\
  Peter Grünberg Institute \\
  Forschungszentrum Jülich \& RWTH Aachen \\
  \texttt{a.bedhotiya@fz-juelich.de}
  \And
  Viet Anh Khoa Tran\\
  Peter Grünberg Institute \\
  Forschungszentrum Jülich \& RWTH Aachen \\
  \texttt{v.tran@fz-juelich.de}
  \AND
  Eleni Nisioti\\
  IT University of Copenhagen\\
  Copenhagen, Denmark \\
  \texttt{enis@itu.dk} \\
  \And
  Sebastian Risi \\
  IT University of Copenhagen\\
  Copenhagen, Denmark \\
  \texttt{sebr@itu.dk} \\
}

\begin{document}
\newcommand{\refeq}[1]{Eq.~(\ref{#1})}
\newcommand{\refSI}[1]{SI Sec. \ref{#1}}
\newcommand{\refsec}[1]{Sec. \ref{#1}}
\newcommand{\reftab}[1]{Tab.~(\ref{#1})}
\newcommand{\reffig}[1]{Fig.~(\ref{#1})}

\maketitle

\begin{abstract}
		Continual learning is the study of how a model accumulates knowledge from data received over time, rather than all at once. In practice, this is typically achieved by dividing a dataset into tasks and presenting them sequentially. Explanations of the resulting measurements take into account factors such as the similarity between segments, the rate at which the data changes and the order of presentation. Each of these factors can be defined in terms of the underlying data, the partition imposed on it, or the learner that encounters it. These levels do not necessarily coincide, and it is usually left implicit which level a reported quantity refers to. We separate these and identify which properties acquire a value at each level. In controlled teacher--student experiments, we demonstrate that partitioning the same underlying trajectory at different levels of granularity, changing the order of presentation, and varying the similarity structure of the underlying data all result in different measured outcomes. We review representative benchmarks to determine which of these properties are stated. Where the underlying data cannot be inspected, as in foundation models built on undisclosed pre-training corpora, its properties are inherited by the model's initial state, and the training applied on top of it can be described.

        
\end{abstract}

%
%

\section{Introduction}\label{sec:intro}

\begin{figure}[t]\centering\includegraphics[width=\linewidth]{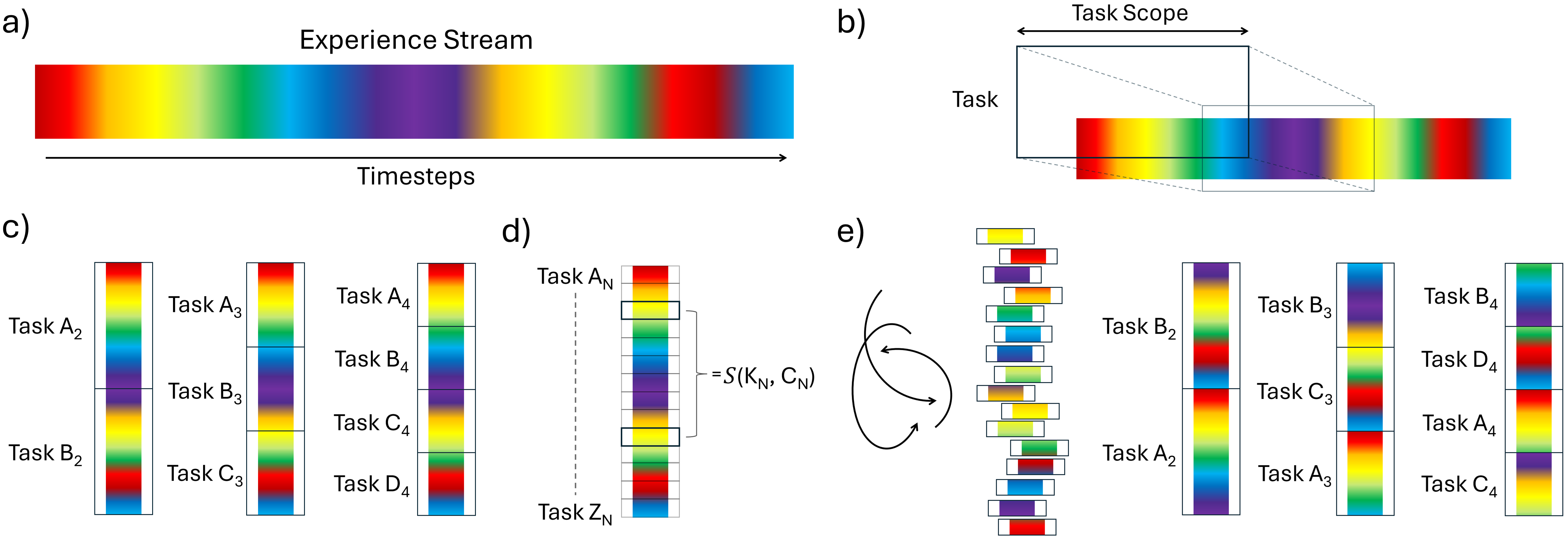}
        \caption{\textbf{One stream, many partitions.}
        \textbf{(a)} A stream is a realized, ordered sequence. Colour denotes content of experience; nearby
            colours indicate related experience.
        \textbf{(b)} A task cut produces a task sequence. Segment width = task scope.
        \textbf{(c)} Change within a task depends on scope.
        \textbf{(d)} Similarity between tasks depends on scope: coarse segments span heterogeneous stretches and overlap; fine segments are narrow and largely disjoint. $S(K_N, C_N)$ is the similarity of two tasks under a partition into $N$ tasks. 
        \textbf{(e)} The same segments in a different order are a different stream, since a stream is a realized order.
        None of (c)-(e) is a property of the stream alone; each is dependent on stream \& cut. Everything here is learner independent. Fig.~\ref{fig:interaction} shows what changes once a learner is present.
        }
        \vspace{-10pt}
        \label{fig:partitions}
        \end{figure}

        Continual learning (CL) refers to a model’s ability to update its knowledge and skills as new data arrives over time, as well as retaining knowledge from past data \citep{van2022three, van_de_Ven_2025}.
        Performance metrics, such as transfer and forgetting, are evaluated on a model trained across a data or experience stream, which is a realized ordered sequence of functions to be learnt (\reffig{fig:partitions}(a)). 
        Typically, this stream is split into disjoint parts, tasks, and the learner is sequentially permitted access to only one task at a time during training.
        Transfer then measures how the model performs as it transitions from task to task, and forgetting measures how well it performs on previous tasks as it learns newer ones.
        
        This standard approach requires deciding how, or whether, to define a \textit{task cut}, the partition of a data stream into separate tasks.
        In practice the strategies for laying boundaries between tasks vary widely.
        In classification scenarios, tasks can be disjoint subsets of classes to be learnt \citep{zenke2017continual, rebuffi2017icarl} or different input transformations with a shared label space \citep{goodfellow2013empirical}, while reinforcement learning typically defines tasks as new goals or skills to be acquired by an agent \citep{wolczyk2021continual, liu2023libero}.
        Alternatively, 
        streaming benchmarks vary the distribution of data to be modeled continuously, and either leave the task cut undefined \citep{cai2021online, verwimp2023clad} or impose it by time period \citep{lin2021clear}.
        Wherever drawn, the cut fixes how much experience falls within each task and how much the learnt function changes inside it (\reffig{fig:partitions}(b,c)), how similar the resulting tasks are to one another (\reffig{fig:partitions}d), and the order in which they are presented, which need not follow the stream's own progression (\reffig{fig:partitions}(e)). 

        Classical CL taxonomies do not account for the task cut, instead differentiating between scenarios based on the availability of task identity during training and testing as well as whether the label space is shared or expanded across tasks \citep{van2022three}.
        These differences have guided most experimental approaches to CL over the past years, yielding many insightful and practically useful results.
        However, it has become increasingly clear that the task cut cannot be ignored when attempting to understand forgetting and transfer in CL models as changing the boundaries of tasks changes the outcome \citep{filat2026temporal}. A stream and a cut are in turn meaningless without a learner attempting to model the data presented.
        Here again, scope, or the history a CL learner has access to, similarity, now within the representational space of the learner rather than the functional space of the stream, and change, in terms of the update rules applied to the learner, are critical to the final forgetting or transfer exhibited.  
            
        In summary, a (final) CL setting is a realized stream, a task cut on it, and a learner moving through the stream. 
        Notions such as scope, similarity, and change are defined differently at these three levels and affect forgetting and transfer differently.
        Benchmarks usually define settings near the stream (classes, domains, transforms). 
        However, much of the theory that predicts forgetting quantitatively works near the learner: NTK overlap \citep{doan2021theoretical}, representational similarity \citep{ramasesh2020anatomy}, feature/readout decomposition \citep{lee2021continual, hiratani2024disentangling}, node re-use vs activation \citep{lee2022maslow}, representational geometry \citep{menghi2025effects, johnston2024modular}.
        
        \textbf{Contributions.} In this work, we describe critical properties of the three levels at which CL operates and provide a framework by which to compare different benchmarks and experimental paradigms. Specifically we:\\
        \textbf{a)} Separate and conceptualize the properties such as scope, order, similarity and change at the stream, task and learner levels (\refsec{sec:theory}, \reftab{tab:property-levels}).\\
        \textbf{b)} Show that in a controlled teacher--student setting partitioning the trajectory differently, reordering the resulting tasks, and varying the distance the stream traverses each change measured forgetting (\refsec{sec:theory}, \reffig{fig:interaction}).\\
        \textbf{c)} Derive the consequence for attribution and for measurement metrics themselves, which are defined over segments (\refsec{sec:attribution})\\
        \textbf{d)} Review representative benchmarks against these properties, showing which are specified, which are left implicit, and where the omissions make reported comparisons unsupported (\refsec{sec:audit}).
        

        We do not claim a complete description and formalization of the listed properties. Extending the experiments to standard benchmarks and to learner-level properties, and specifying and testing the consequences we draw for foundation models, are left to future work. Complexity and impossibility results exist for the continual learning problem as usually posed \citep{knoblauch2020optimal, peng2022continual}, taking the partition as given. Defining tasks is also contested outside ML \citep{Gozli2019-GOZWIA}.

\section{Continual Learning: The three levels}\label{sec:theory}

    Prior CL work has explored numerous properties of how the data is presented to the learner such as order, similarity, or rate of change (or drift).
    These analyses are often conducted on a predetermined partition of the dataset, and hence fail to distinguish whether the conclusions hold for the actual data stream or merely for the selected partition.
    Furthermore, a subset of literature also demonstrates these properties to be heavily dependent on the actual learning rule and underlying hypothesis space~\citep{chen2020mitigating,WANG2025168}.
    Outside of the task-dependent framework the literature investigating these properties is scattered across different sub-fields of CL. Therefore, we introduce a three-level taxonomy -- \textit{stream}, \textit{task}, and \textit{learner}, with the distinction of the properties shown in  (Table~\ref{tab:property-levels}). 
    
    In this section, we introduce our taxonomy, re-contextualizing related work and demonstrating practical implications of our framework in a simplified toy experimental setting.
    Inspired by the student-teacher setup used in \cite{lee2021continual}, we train a 2-layer ReLU MLP, the student, to reproduce the outputs of a smaller two-layer ReLU MLP, the teacher.
    To specify a continual task stream, we gradually shift the weights of the teacher MLP along a smooth trajectory that can be specified arbitrarily through a series of linearly interpolated way-points in weight space.
    Since local shifts in weight space result in locally bounded changes to the teacher's function, this simulates a realistic continuously shifting task stream while also enabling control over the stream's scope, resolution and order.
    For simplicity, we use only plain Adam optimization on the student, focusing these experiments mostly on how task cuts and modifying properties affect what the learner retains across the stream in a simple student network. 
    Details regarding the implementation of every experiment can be found in Appendix~\ref{app:toy}, with results shown in Fig.~\ref{fig:interaction}(c)-(e) and Fig.~\ref{fig:appendix_experiments}.

\begin{table}[ht!]
\small
  \centering
  \setlength{\tabcolsep}{4pt}
  \renewcommand{\arraystretch}{1.25}
  \caption{Properties of the realized stream, task discretization, and
    learner.
}
  \label{tab:property-levels}
  \begin{tabularx}{\linewidth}{
    @{}
    >{\raggedright\arraybackslash\bfseries}p{0.12\linewidth}
    Y{0.90}
    Y{0.95}
    Y{1.15}
    @{}
  }
    \toprule
    Property
      & \textbf{Stream} (\S\ref{sec:stream})
      & \textbf{Task Cut} (\S\ref{sec:cut})
      & \textbf{Learner} (\S\ref{sec:learner}) \\
    \midrule

    Scope
      & Support of the realized stream (data)
      & Support of a task/segment induced by the partition\par\smallskip
      & Support of the learner's trajectory in function space.
        \par\smallskip
         Access: the learner's filtration \\
    \midrule

    Order
      & Realized order at sample or batch level
      & Order of the task labels induced by the partition
      & Which learner state encounters which segment \\
    \midrule

    Similarity
      & A function of time points, \(D(t,t')\), or an aggregate
      & Intra- and inter-task similarity
      & Distance in hypothesis or representation space \\
    \midrule

    Change
      & Gradual change or drift over time
      & Discrete change across task boundaries, and intra-task change induced by partition
      & State displacement or learning dynamics induced by the
        update rule \\
    \bottomrule
  \end{tabularx}
\end{table}

        \subsection{Stream}\label{sec:stream}
        Learning is sequential for a non-methodological reason because a learner encounters experience in one order and cannot revisit it except through what it has retained in its parameters, in context, or in an explicit buffer \citep{wang2024comprehensive}.
        We use the term \emph{stream} for a realized, ordered sequence, not the distribution it could have been drawn from~(Fig.~\ref{fig:partitions}(a)).
        Many regimes are specified distributionally, as multiple epochs over a dataset, or repeated draws from a task distribution. However, the learner only encounters this distribution through this realized stream and the final outcome depends on the order of data which has been available. 

        \textbf{Scope of the realized stream.} The scope describes the stream's data support, i.e.\ the breadth of distinct experience the stream covers as a whole. Benchmarks name the source a stream is drawn from, but no standard measure of its diversity exists, so the scope is bounded without being quantified, while influencing the lower and upper bounds of transfer and forgetting. In empirical CL literature, scope ranges from a single digit dataset (e.g. Split-MNIST) to a mixture of generalist and specialized datasets as in continual pre-training of foundational models~\citep{ibrahim2024simple}. Our teacher-student setup allows for changing the scope of the task stream by increasing the intrinsic dimensionality across which the weights of the teacher move through the stream. 
        Changes in stream scope affecting the learner's retention are demonstrated in Appendix~\ref{app:toy}.
        
        \textbf{Realized order.} The order describes which experience precedes which (e.g., the ordering of samples or batches). Order arises from the environment's dynamics in natural settings, from the learner's own behavior in embodied and interactive settings, and from the experimenter in benchmarks.
        Stream-level order is rarely remarked on by CL literature, but rather accounted for implicitly, aggregating statistics over multiple random number generator seeds, inducing different data point orderings. In our teacher-student setup, we can switch the order of a task stream by traversing it in the forward or reverse direction. Results to this effect (along with results regarding a shuffled presentation once a task cut is decided) are shown in Fig.~\ref{fig:interaction}(d). Other changes to how waypoints in the teacher's weight space are traversed would necessarily alter the underlying nature of the stream itself.
        
        \textbf{Similarity.} Stream-level similarity measures how different one sample or region is from another, ignoring the observer and learner. This requires a metric on experience itself, which can then be used to make comparisons. The choice of space this metric lives in (parameter, data, function) needs to be further given, and \citet{mitchell2025beyond} note that it is sometimes forced, as when differing architectures rule out parameter-space distance. \citet{webb2016characterizing} make the abstraction explicit, defining drift against an unspecified distance function and suggesting Hellinger or KL as instances.
        These measures are often asymmetric, how informative one region is about another need not equal the reverse.

        \textbf{Change.} The temporal geometry of the similarity can be abrupt, gradual, cyclic, compositional and not necessarily evenly distributed along the stream. In concept drift literature, this has been described as the \textit{drift rate} at some time step $t$~\citep{webb2016characterizing}. The effective learning regime depends jointly on change and similarity \citep{graldi2025importance}. 
        In our setup, we are able to manipulate the overall similarity along with the rate of change within the stream by increasing the maximum distance that is traversed by the teacher weights along the stream trajectory. Fig.~\ref{fig:interaction}(e) demonstrates that increasing this distance, equivalent to decreasing the similarity of any region in the task stream to any other region, increases the final test loss across all tasks, our measure of retained performance.


        \subsection{Partitioning and tasks }\label{sec:cut}
        A stream comes with no predetermined notion of tasks or boundaries. In fully online settings none are required, since the properties above describe the stream as a whole. However, partitioning the stream into segments can often be indispensable for prediction and evaluation. 
        In benchmark construction the choice is explicit. Two MNIST digits are one such segment, the next two another. In natural settings however there is no innate partition, and segments a researcher describes are a retrospective device.
        Task-free CL methods can automatically locate boundaries \citep{aljundi2019task, van_de_Ven_2025}, but still require a distinction criterion supplied by the experimenter, with different criteria locating different boundaries in the same underlying stream.
        There is also no privileged granularity: locomotion, running, or extending a leg are defensible units for the same behaviour. 
        Segments carrying the same label form an equivalence class we call a \emph{task}. The label is not given by the stream and requires either an external (annotation, class partition, designed task set) or internal (measured similarity under a threshold) criterion. 
        The \emph{scope} of a task is the support of the subset of the realized stream associated with the task (Fig.~\ref{fig:partitions}(b)).         

        \textbf{Task similarity and change.} With the task cuts determined, inter-task similarity becomes well-defined and measurable by distributional~\citep{alvarezmelis2020geometric, liu2025wasserstein} or functional distance~\citep{lee2021continual,hiratani2024disentangling}. 
        Different granularities of partitioning render different inter-task similarities and change (Fig.~\ref{fig:partitions}(c,d)).
        This model-agnostic notion should not be conflated with the model-dependent similarity discussed in Section~\ref{sec:learner}.
        
         \textbf{Task order.} Once tasks exist, the order of their instances can be manipulated, yielding a different stream (Fig.~\ref{fig:partitions}(e)). \emph{Inter-task} order is how instances of different tasks interleave. \emph{Intra-task} order or \textit{exposure} is how instances of one task are distributed across the stream, massed, spaced or recurring. Blocked (contiguous per task) and interleaved (tasks alternate) are two points on this axis, which also includes spacing and number of traversals. Prior literature has already documented the effect of reordering ~\citep{bell2022effect,mitchell2025beyond}, and of task similarity ~\citep{lee2021continual,hiratani2024disentangling} on forgetting. However, order and task similarity cannot be cleanly manipulated independently from each other. \citet{ramasesh2020anatomy} demonstrate that the preceding tasks determine whether high inter-task similarity leads to more or less forgetting. \citet{lin2023theory} analytically describe the expected forgetting and generalization error as a function of both similarity and order, while
        \citet{li2025optimal} provide an analytical reordering scheme based on the similarity structure.
        Task similarity also affects the performance of interleaved and blocked presentations, coinciding with human learning behavior~\citep{menghi2025impact}.

        In our setup, we vary the granularity of the task partition on a fixed teacher trajectory under a constant compute budget, effectively changing both inter-task similarities and intra-task change.
        Fig.~\ref{fig:interaction}(c) shows that simply discretizing the task stream more finely can alter the profile of retained loss across the stream, such that more dissimilar regions to the final task exhibit a higher test loss after training while more similar ones (even if they occur much earlier in the task stream) are preserved better than under coarser discretizations. Fig.~\ref{fig:interaction}(d) further investigates the effect of task order, revealing again a dependence of the shape of forgetting across the task stream on the order in which tasks were presented. Randomly shuffling the order removes any relation between a region's position in the stream and how well it is retained. In contrast, reversing the order of task presentation clearly reverses the shape of forgetting across regions of the stream.
        
        \subsection{The learner}\label{sec:learner}
        
\begin{figure}[t]\centering\includegraphics[width=\linewidth]{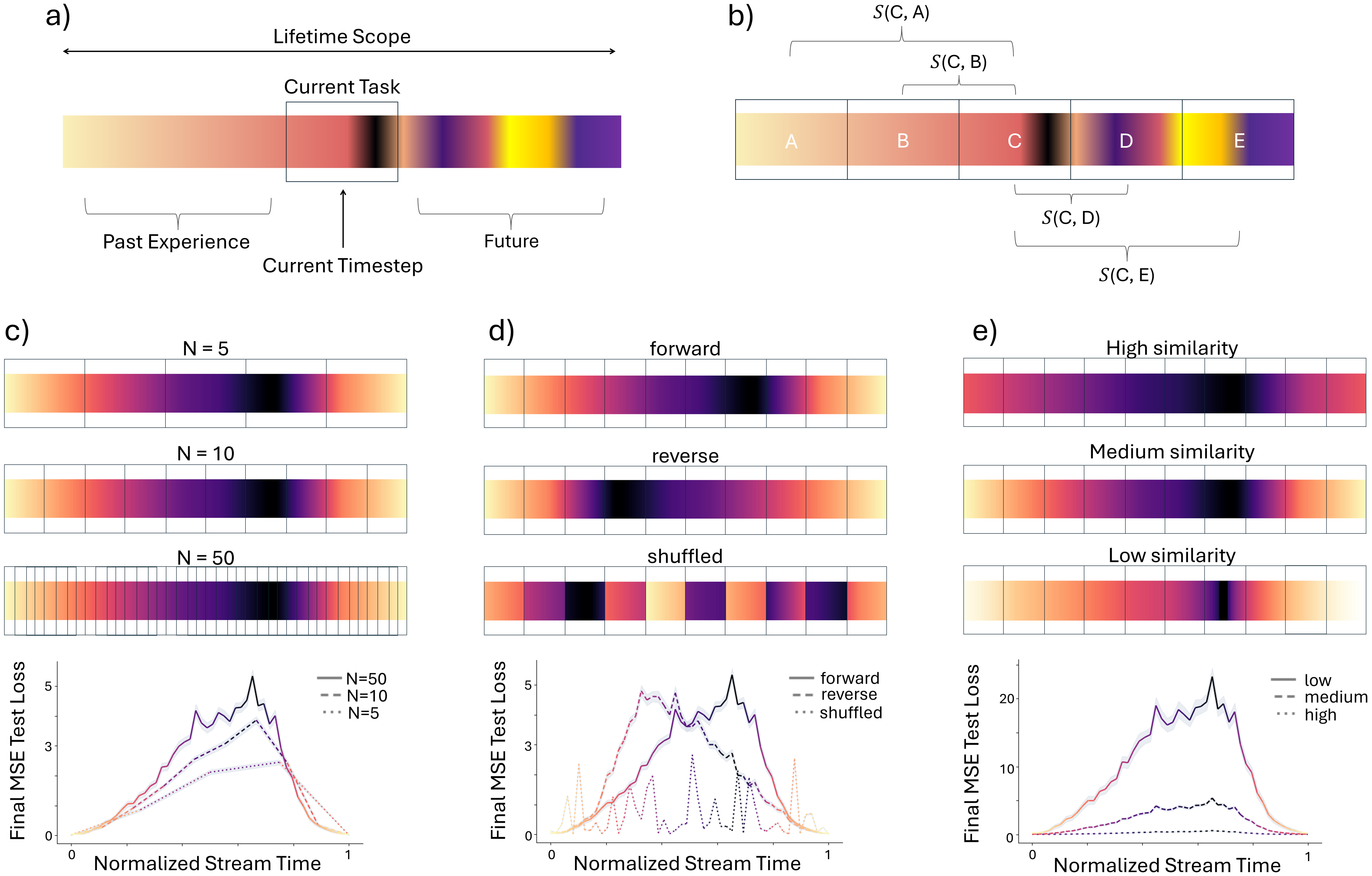}
        \caption{\textbf{A learner enters the setting.}
        \textbf{(a)} A learner occupies one point in a stream at a time. Its lifetime scope is the breadth of experience the stream covers as a whole, of which only the past is available to it and the future is unobserved.
        \textbf{(b)} The same stream under a single partition. Similarity is defined for every pair of tasks and computable from the data once the boundaries are fixed. What a learner retains of each task is a separate quantity (\refsec{sec:learner}).
        \textbf{(c--e)} A teacher--student stream in which the teacher's weights follow a fixed trajectory, with the student trained on tasks sampled along it. Curves show mean squared test error on each task after the whole stream has been traversed, shaded by standard error over five initialisations, with task position on the horizontal axis. Full setup in \refsec{app:toy}.
        \textbf{(c)} Partition granularity, $N \in \{5, 10, 50\}$, which alters similarity between tasks and change within them together.
        \textbf{(d)} Presentation order at $N = 50$. Reversing the stream reverses the profile of retained error across it, and shuffling removes its structure.
        \textbf{(e)} Distance traversed by the teacher across the stream at $N = 50$, which sets the overall similarity between its regions.
        }\label{fig:interaction}\end{figure}

        With a learning agent in the picture, the relevant domain shifts from data to the hypothesis space \citep{vapnik} where the stream carves a learning trajectory, an ordered distribution over this space. New concepts emerge, such as capacity, inductive biases, architecture, optimizer, optional pretrained representations, and the choice of updated component (e.g., in-weight vs in-context adaptation \citep{dorovatas2026position}, drifting representation vs stable readout \citep{vanderveldt2025drift}).
        Since each decomposition is built for different purposes, they do not easily translate across domains, e.g., in language the representation and the task-solving component are not separable as in vision. 
        As with the stream and task level, we can map scope, order, similarity, and change to the learner:

        \textbf{Scope and access.} The learner's scope is the support of the stream induced trajectory (\reffig{fig:interaction}). The learner's access includes the accessible stream history (filtration) of the stream and possibly cuts/task identities until the present time. Prior literature categorizes the accessibility to task identity as task, class and domain-incremental setups~\citep{hsu2019reevaluatingcontinuallearningscenarios,van2022three}.
        For learners accessing the task identity, the task support is consequential: for instance, \citet{prabhu2024randumb} report continual prompt-tuning methods collapse as the number of classes per task is reduced to one.
        
        \textbf{Agent-side similarity, order and change.} Similarity for the agent refers to model and trajectory dependent measures like representational or NTK overlap \citep{doan2021theoretical, ramasesh2020anatomy}. This combines the agent properties with the task cut, therefore cannot be reported independently of both. Can also be viewed as distributional distance on the function space induced by a metric e.g., gradient or parameter overlap. 
        The \emph{update rule} specifies the change in the trajectory in response to the stream and possibly the cuts. 
        Under \emph{replay}, samples retained from earlier segments are re-inserted into the sequence the update rule receives, so the realized stream of \refsec{sec:stream} and the data the learner trains on come apart.        
        Two settings that share a stream and a cut, but present the segments in a different order, place each segment against a different accumulated learner state.

        
        \textbf{Effective capacity and difficulty.} The regime in which the learner is operating is a consequence of the stream's demands relative to the capacity. This can be a result of parameter count relative to the stream or due to hard (scope restriction) or soft (modifying tails) inductive biases. Under a learning strategy that uses boundaries only for evaluation, this is fixed by stream and architecture alone. 
        As an example, in overparameterised settings the network accommodates the entire stream, so no region of it strains capacity. What then varies across conditions is not the stream but the task construction, and it is that similarity the forgetting results are reported against \citep{ramasesh2020anatomy, lee2021continual, lee2022maslow, hiratani2024disentangling}. A related composite of effective capacity and dataset is \emph{difficulty}, conceptualizing how well the task can be represented in the model's function space. Concretely, the  minimum KL divergence between the optimum likelihood and the true task distribution. The same task can be trivial for one learner and hard for another, which is why difficulty cannot be reported by only a benchmark \citep{tang2025importancetaskcomplexityevaluating}. 
        
      
        A crucial metric that allows the learner to optimise its strategy is \emph{predictability}, the ability to anticipate what comes next: the future trajectory bounded by the stream's properties, but realized only relative to what this learner has retained and can potentially extrapolate from. Predictability determines the worth of retention as where the future is unrelated to the past, retention can cost capacity without benefit, where it is closely related retention is directly helpful. \reffig{fig:interaction}(a) shows the sequential nature of learning, standing at one point in the stream with past behind and future unobserved.

\subsection{The attribution consequence}\label{sec:attribution}


        What follows is that an outcome in CL is a joint function of the setting and the learner $agent$.
        \[
        \text{outcome}=f\big(\underbrace{\text{stream},\ \text{cut},\ \text{access}}_{\text{setting}};\ \text{agent}_t\big),
        \qquad
        \text{agent}_t=g\big(\text{agent}_0,\ \text{stream}_{<t},\ \text{access}_{<t}\big).
        \]
        The setting therefore matters directly, as the conditions of measurement, and indirectly, through the learner those conditions produced. Two studies sharing a stream and cut but ordering segments differently are evaluating two different learners. Even when holding the method fixed, results are not directly comparable across settings, so a statement like "method A beats method B" compares learners whose difference is not exhausted by A versus B. For example, \citet{ibrahim2024simple} show that what they define as a stronger shift (Pile$\to$German Common Crawl) requires more \emph{replay} than what they define as a weaker one (Pile$\to$SlimPajama) to mitigate forgetting to the same extent, so a reported percentage bundles a stream property with the learner's objective and accumulated state.

        Forgetting, backward and forward transfer are computed from an accuracy matrix whose indices are the cut. The consequence of this is that per-sample loss and accuracy change with the cut only if the learning strategy is affected.
        This predicts that where the update rule does not see the cut, loss and accuracy should be invariant while forgetting and transfer are not, since the latter are defined over segments chosen after training.
        Boundaries are therefore necessary for metrics even where they are not necessary for training. In a fully online setting they are not a training requirement. Standard CL scenarios differ in how they access the cut: \emph{Single-head online SGD} only sees the cuts as annotations. \emph{Multi-head task-IL} uses the cut to select the active head and routes gradients by it. \emph{Class-IL} fixes the output space by it. \emph{Boundary-triggered operations} like EWC Fisher computation \citep{kirkpatrick2017overcoming} and LwF teacher snapshot \citep{LwF2018} fire on it, changing the objective at the boundary. \emph{Replay} is the strongest case, since the cut decides what enters and leaves the buffer and therefore which data the learner sees at all. This yields the prediction that where two cuts induce the same schedule and evaluation reference, results should agree. Consistent with this, \citet{filat2026temporal} hold stream, model and budget fixed and see forgetting and backward transfer move with the split.

        Given a fixed stream, cuts can be chosen to favor retention, plasticity, or whatever the objective demands. Task construction is consequential even where the construction also alters what each task contains: \citet{jhajj2025elastic} report roughly 9.8 percentage points more forgetting under relation-based than under random partitioning of a knowledge graph, and conclude independently that protocols should report their partitioning strategy. There is no existing selection procedure for the best way to partition tasks. \citet{filat2026temporal} propose a measure of how sensitive a stream is to where its boundaries fall, but do not use it to choose them, and \citet{li2025optimal, bell2022effect} optimise order with the task set already given.
        Any such procedure depends on the objective, a cut minimising interference does not have to be the same as one maximising transfer. 

        \newpage
        
\section{What Benchmarks Record}\label{sec:audit}

        We reviewed CL benchmarks against the properties of \refsec{sec:theory}, recording for each property whether the benchmark's construction specifies it (\refsec{app:audit}). Benchmarks built by splitting a classification dataset into class subsets, or by transforming its inputs, are inexpensive, reproducible and comparable across papers \citep{zenke2017continual, rebuffi2017icarl, goodfellow2013empirical}. 
        Benchmarks specify a procedure for constructing a stream rather than a realized stream, holding many factors fixed by construction, which is what makes them controlled instruments. Much of what is reliably known about forgetting and transfer was obtained with them. Which data are drawn on is fixed, but in synthetic benchmarks the assignment of classes to tasks and the order in which tasks are presented are commonly drawn per run and averaged over seeds, so two studies reporting results on Split-CIFAR-100 need not have trained on the same task sequence. Benchmarks built from temporal data fix the order instead \citep{lin2021clear}.
        Split-CIFAR-100 is run with anywhere from 2 to 50 classes per task \citep{rebuffi2017icarl, van2022three}, and its standard random split does not state how closely the classes within a task resemble one another. A method performing well on it may be performing well on closely related or on unrelated tasks.
        In line with these observations, \citet{verwimp2023clad} survey benchmarks and conclude that most represent extreme rather than typical CL problems; \citet{lin2021clear} report that protocols training and testing on i.i.d.\ data inflate measured performance; \citet{jhajj2025elastic} find that how tasks are constructed affects measured difficulty and recommend that protocols state their partitioning strategy. 
          
        \textbf{Access.} Whether task identity is supplied, at which stage, and whether transitions between tasks are signaled is fixed by every benchmark we reviewed, though not always stated. The three often cited standard scenarios \citep{van2022three} focus on task identity at test time, and class-incremental learning additionally requires discrimination across all classes seen so far. For Permuted MNIST, the scenario follows only from the output layer shared across tasks, which \citet{van2022three} classify as domain-incremental.
        
        \textbf{Similarity and scope.} Split benchmarks state how classes are grouped, while permutation benchmarks draw a random input permutation for each task \citep{kirkpatrick2017overcoming}, which can therefore differ between studies.
        While a reader can infer that some pairs of tasks are more closely related than others, they cannot determine by how much or in which respect, since inputs may overlap where labels do not. Where specified, \citet{liu2023libero} declares which axis shifts, and \citet{ibrahim2024simple} additionally grade their shifts as weak or strong. The only quantitative measure, forward transfer between every pair of tasks \citep{wolczyk2021continual}, is defined for a reference agent and therefore lies at the learner level (\refsec{sec:learner}).
        Scope is conveyed similarly, where the number of classes per task is known but not comparable across benchmarks. The exception among the reviewed is \citet{liu2023libero}, which declares per suite which factors vary, though without a magnitude.
        An instruction-tuning dataset \citep{wang2023trace} and a discrimination between two handwritten digits \citep{zenke2017continual} are each called one task.
        
        \textbf{Change and exposure.} Among the benchmarks we reviewed, these two properties are not varied independently. Most benchmarks with abrupt transitions present experience in blocks that can be revisited. The other combinations occur in \citet{lin2021clear} and \citet{yao2022wild}, which present gradually drifting data in blocks, while CLAD-C and CL-Bench contain sudden shifts within an online stream \citep{verwimp2023clad, asawa2026continual}. No reviewed benchmark varies one property while holding the other fixed on the same stream, so that an outcome cannot be attributed to either alone.
        
        To conclude, comparing methods on a fixed benchmark does reveal differences in how learners behave. However, what such a comparison establishes is local to that benchmark, and to generalize the behavior, a reader would need values for all properties on which the settings differ.

\section{Foundation Models and Beyond}\label{sec:fm}

       
        In foundation models the properties of \refsec{sec:theory} are harder to report, since the division into tasks may have no counterpart in how the model was trained. Surveys divide CL for foundation models into \textit{stages}, contiguous training periods under one objective and one data source: continual pre-training, domain-adaptive pre-training, continual fine-tuning and alignment \citep{shi2025continual, yang2025recent}, or continual pre-training, fine-tuning and compositionality \citep{bell2025future}. Naming them mirrors the operation of \refsec{sec:cut}, with the surveys positioning the divisions differently. The stages mix two criteria: continual and domain-adaptive pre-training change the data source, which is a cut in the sense of \refsec{sec:cut}, while alignment changes the objective and therefore the update rule, which \reftab{tab:property-levels} places at the learner level. Forgetting reported across a stage boundary confounds a cut-level with a learner-level change.
        Alongside stages, \citet{shi2025continual} classify shifts during continual pre-training as temporal, content-level or language-level. A temporal shift may span a month or a decade and a content shift may cover adjacent domains or unrelated ones, so two studies reporting the same type of shift may have faced very different problems.         
        Two settings have to be separated. Where continual pre-training is itself the object of study, experimenters choose the corpora, their mixture and order \citep{ibrahim2024simple}, stream-level properties remain reportable, and the constraint of a limited replay buffer is replaced by compute budget, since i.i.d.\ mixing reduces the throughput of novel data
        \citep{prabhu2023onlinecontinuallearningstorage}. Where a released checkpoint is the starting point, as in continued multilingual pre-training \citep{rei2026tower} or evaluation of deployed models \citep{wang2023trace}, the pre-training corpus is usually unpublished \citep{wan20252025}, and where corpus and order are released the order is a shuffle \citep{biderman2023pythia}, so even fully disclosed data carry no meaningful stream-level order.
        Here the stream-level properties are fixed before the study begins and enter it through $\text{agent}_0$. The adaptation stream and its cut are specifiable, while the scope of experience the checkpoint covers remains a black-box assumption on the learner state, on which any reported forgetting is conditioned (\refsec{sec:attribution}). Reusing the checkpoint holds that scope fixed across studies, which makes agent-side similarity, defined relative to a learner and a cut (\refsec{sec:learner}), comparable between them.
        \citet{ibrahim2024simple} match full re-training when a new corpus arrives as a mixture of domains but report poor results when the same corpus is consumed one domain at a time, and find that the stronger shift requires more replay than the one they classify as weaker. Continued multilingual pre-training still degrades general-purpose benchmarks, requiring further alignment stages \citep{rei2026tower}. Proposals for the field's future lie at the learner level, whether
        compositionality, the recombination of components \citep{bell2025future}, or memory operations deciding what to store, when to consolidate or retrieve and what to discard \citep{dorovatas2026position}. \citet{asawa2026continual} report naive in-context learning outperforming systems built around such operations on tasks whose instances are constructed to share a latent structure. What a memory policy should store depends on the predictability of the stream it runs on (\refsec{sec:learner}).

        \textbf{Outlook.} Any system that keeps learning after an initial training phase learns continually. As a learner is a product of the setting in which it was trained (\refsec{sec:attribution}), attributing a performance difference to a method requires knowledge of that setting, which is why we ask for the properties of \refsec{sec:theory} to be routinely reported. Together they describe a space of learning problems. A robot restricted to loading a dishwasher in a kitchen and a model deployed across changing domains both learn continually, over different scopes of experience. What a learner should retain and what it should revise depends on how that experience is structured, and the relation is not necessarily monotonic \citep{ramasesh2020anatomy, lee2022maslow, hiratani2024disentangling}. Such a description would also help the learner to estimate its lifetime scope, to then decide what and how much to retain. Revising that estimate changes which experiences it treats as similar and how its capacity is best spent. In interactive and agentic 
        settings the stream is generated in part by the learner's own behavior, so setting and learner are jointly determined rather than composed as in \refsec{sec:attribution}. Scope becomes a property of the policy, and two methods evaluated on the same agentic benchmark traverse different streams. The field lacks a means of predicting expected forgetting and transfer for a given stream, cut and learner. Currently, a replay fraction that works on one corpus pair, or a method that performs well on one split of CIFAR, gives little generalisation. \citet{hinder2020towards} formalise how a data distribution evolves over time, which supplies the stream alone. \citet{mitchell2025beyond} propose a different approach, algorithmic invariance to permutations that carry no information; we consider the two approaches complementary as identifying which permutations these are requires the specification we request.
        
\section*{Acknowledgements}
This project was supported by a European Research Council (ERC) grant (GA no. 101045094, project "GROW-AI"), and by Helmholtz Association's project-oriented funding programme (PoF 2, Topic 3). Aspects of this project originated from interactive discussions and hands-on working groups at the CapoCaccia Neuromorphic Intelligence Workshop (https://capocaccia.cc), which the authors gratefully acknowledge.


    \bibliographystyle{plainnat}
    \bibliography{references}


\appendix

\section{Benchmark review: full entries}\label{app:audit}

We selected benchmarks that are widely used in CL and that together span classical synthetic partitions, purpose-built streams, interactive settings and foundation-model training. The selection is illustrative rather than exhaustive, and the entries below describe each benchmark's standard protocol.

    \begin{table}[ht]
            \centering
            \caption{\textbf{Benchmark families against the properties of \refsec{sec:theory}.}
    Superscripts denote the \emph{status} of the entry (\spec\ specified; \est\ not stated by the benchmark but our reading of its description, judged recoverable from the released data and not computed here; \unk\ unspecified and not recoverable from the protocol, with a marker alone indicating that the protocol gives no value). For Similarity, \spec\ marks that the benchmark states how tasks were constructed, which constrains the kind of relatedness without providing a magnitude. Two of the reviewed benchmarks provide more information. \citet{ibrahim2024simple} grade their two shifts as weak or strong, and \citet{wolczyk2021continual} report forward transfer between every pair of tasks for a reference agent, a learner-level quantity that they also use to set the task order. For Scope, a class count fixes how many labels fall in a task without stating the breadth of experience they cover, and is recorded as \unk; only \citet{liu2023libero} is marked with \spec, for declaring per suite which factors are held fixed and which vary. The Order/Exposure column records the presentation regime, while the realized ordering within it is typically drawn per run rather than fixed by the benchmark. The last two columns, Task identity at test time and Boundaries, are two facets of the learner's access. Each entry is recorded at whichever level the benchmark specifies.}
            \label{tab:audit}
            \resizebox{\linewidth}{!}{
    \begin{tabular}{l llll ll}
    \toprule
    Benchmark & Scope & Order/Exposure & Similarity & Change & Task identity & Boundaries \\
    \midrule
    \multicolumn{7}{l}{\emph{Classical synthetic partitions}}\\
    Split MNIST \citep{zenke2017continual, van2022three}
      & 2 classes per task \unk & blocked \spec & label-disjoint, pairs fixed or per run \spec & abrupt \spec
      & given or withheld \spec & given \spec \\
    Split CIFAR-100 \citep{rebuffi2017icarl, van2022three}
      & 2--50 classes per task \unk & blocked \spec & label-disjoint, grouping per run \spec & abrupt \spec
      & given or withheld \spec & given \spec \\
    Permuted MNIST \citep{goodfellow2013empirical,kirkpatrick2017overcoming}
      & full label set \unk & blocked \spec & labels shared, inputs permuted \spec & abrupt \spec
      & none (shared output) \spec & given \spec \\
    \addlinespace
    \multicolumn{7}{l}{\emph{Purpose-built streams}}\\
    CORe50 \citep{lomonaco2017core50}
      & scenario-dependent \unk & shuffled batch order \spec & same objects across sessions \est
      & abrupt at switch \est & none (single head) \spec & batch-marked \spec \\
    CLAD-C \citep{verwimp2023clad}
      & 6 classes \unk & online, chronological \spec & class and domain shift \est
      & gradual $+$ abrupt \spec & none \spec & none \spec \\
    CLEAR \citep{lin2021clear}
      & 11 classes \unk & offline per bucket \spec & concepts persist \est & gradual \spec
      & none \spec & buckets imposed \spec \\
    CLOC \citep{cai2021online}
      & 712 location classes \unk & online, single-pass \spec & not stated \unk & unspecified \unk
      & none \spec & none \spec \\
    Wild-Time \citep{yao2022wild}
      & varies by dataset \unk & set by protocol \spec & not stated \unk & gradual, timestamped \spec
      & none \spec & timestamps \spec \\
    \addlinespace
    \multicolumn{7}{l}{\emph{Interactive / embodied}}\\
    Continual World \citep{wolczyk2021continual}
      & manipulation goal \unk & sequential \spec & shared spaces, pairwise transfer \spec & abrupt at switch \spec
      & given (multi-head) \spec & given \spec \\
    LIBERO \citep{liu2023libero}
      & fixed factors per suite \spec & sequential \spec & declared axis (3 of 4 suites) \spec
      & abrupt at switch \spec & language instruction \spec & given \spec \\
    \addlinespace
    \multicolumn{7}{l}{\emph{Foundation-model era}}\\
    TRACE \citep{wang2023trace}
      & one dataset $=$ one task \unk & blocked, fixed order \spec & heterogeneous \unk
      & abrupt dataset switch \spec & in-context \spec & given \spec \\
    Continual pretraining \citep{ibrahim2024simple}
      & corpus \unk & shuffled, single pass \spec & shift graded weak/strong \spec
      & abrupt corpus switch \spec & none \spec & dataset-marked \spec \\
    Continual SSL \citep{fini2022self}
      & split-dependent \unk & blocked, many epochs \spec & not stated \unk & set by split type \spec
      & setting-dependent \spec & required \spec \\
    CL-Bench \citep{asawa2026continual}
      & domain task \unk & online, episodic \spec & latent structure shared \spec & task-specific shifts \spec
      & none \spec & instance-marked \spec \\
    \bottomrule
    \end{tabular}
            }
        \end{table}


\newpage

\section{Toy experiments}\label{app:toy}



To investigate how continuous task streams and the task cut affect a learner's forgetting, we deploy a simple synthetic teacher-student toy setup. 
In real-world CL settings, data streams are often high-dimensional, complex shifts of data distributions generated by opaque processes. 
This teacher-student setup attempts to model such continuous distribution shifts through explicitly configured, continuous weight trajectories $\Theta(t)$ of a reference teacher network $f_{\Theta(t)}$. 
The teacher $f_{\Theta(t)}$ yields a task stream that can be discretized by sampling target datasets $(X_n, Y_n)$ along the weight trajectories, where properties such as the trajectory distance (reflective of the stream scope), resolution of the discretization (which influences the per-task scope and per-task change), presentation order, and parameter subspace alignment (which allows independent manipulation of the label space, input space or representational space of the teacher network) are controllable and can be evaluated.

\subsection{Teacher architecture and weight trajectory construction}

The teacher network $f_{\Theta(t)}: \mathbb{R}^{10} \to \mathbb{R}^{2}$ is a multi-layer perceptron (MLP) with 2 hidden layers of 32 ReLU neurons.
This results in a total parameter dimension of $D = 1474$. 
Parameters evolve continuously over normalized stream time $t \in [0, 1]$ according to low-dimensional keypoint trajectories $\mathbf{w}(t) \in \mathbb{R}^d$ ($d \ll D$). 
To lift these waypoints into the full teacher weight space $\mathbb{R}^D$, we project them using $P \in \mathbb{R}^{D \times d}$ (with unit-norm columns) sampled from $\mathcal{N}(0, I_D)$, defining the parameter evolution as $\Theta(t) = \Theta_0 + P \mathbf{w}(t)$, where $\Theta_0$ is also sampled from $\mathcal{N}(0, I_D)$.


\subsection{Stream discretization}

To evaluate the impact of discretization coarseness, a continuous teacher weight trajectory is partitioned into $N$ discrete tasks sampled at $t_n = n / (N - 1)$. 
For each task $n$, the teacher $f_{\Theta(t_n)}$ is passed inputs $x_n$ sampled from $\mathcal{N}(0, I_{10})$ and generates outputs $y_n$ which are used as labels to train the learner.
The learner's training data budget across the stream is fixed at $10,000$ samples, allocated evenly across the sampled tasks, so that the total data budget and the total number of gradient steps are constant across discretization levels.
In addition, we use exactly the same underlying weight trajectory for every discretization level (ensured by always using the same random seed during trajectory generation). 
We evaluate discretization levels $N \in \{5, 10, 50\}$, corresponding to $2{,}000$, $1,000$, and $200$ training samples per task. 
For testing, a test set of $200$ samples is independently generated for each task $n$.

Figure~\ref{fig:interaction}(c) shows the results of changing the task discretization averaged over student networks generated with 5 different seeds.


\subsection{Task presentation order}

For a fixed discretization of $N = 50$ and using the same training and test setup as in the previous section, we evaluate three task presentation orders: 
\begin{itemize}
    \item Forward: The same sequence as the stream trajectory $n = 0, 1, \dots, N-1$.
    \item Reverse: The inverted trajectory sequence $n = N-1, N-2, \dots, 0$.
    \item Shuffled: A random permutation of tasks.
\end{itemize}

Figure~\ref{fig:interaction}(d) covers the results from this experiment, again over 5 different seeds for the student network.

\begin{figure} \centering \includegraphics[width=\linewidth]{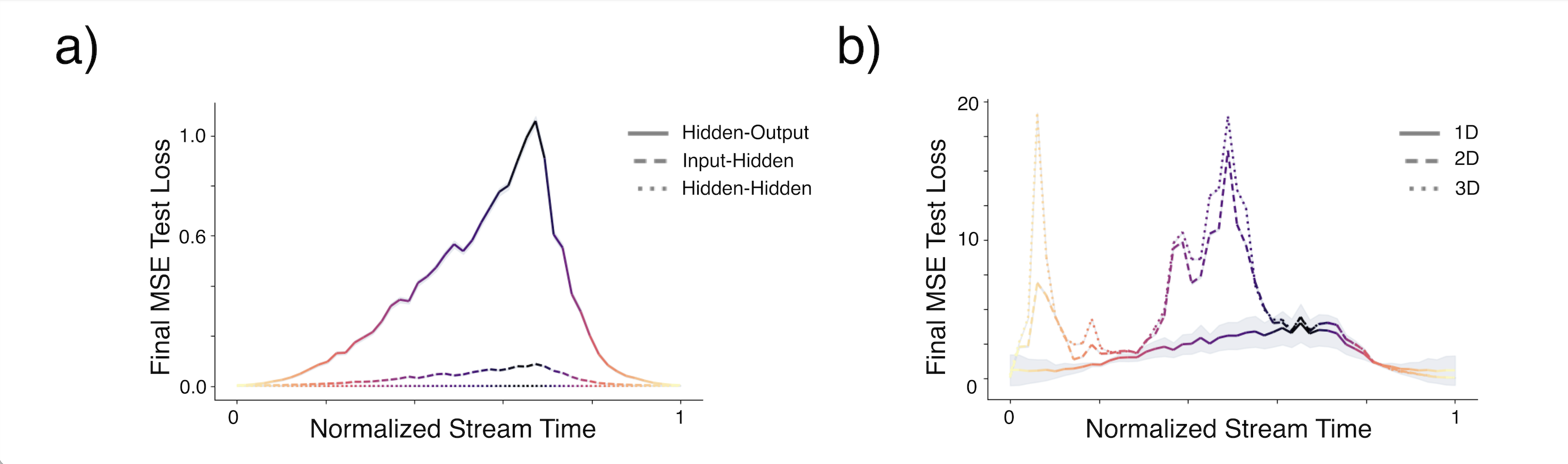} \caption{(a) Changing different parameters in the teacher has different effects on forgetting in the student. (b) Changing the stream scope (i.e., the intrinsic dimensionality of the teacher's weight trajectory) affects forgetting in the learner. } \label{fig:appendix_experiments} \end{figure}

\subsection{Selective teacher shift}

To determine which components of the teacher's function most affect forgetting in the student, we selectively restrict the parameter trajectory shifts to specific weight layers in the teacher while freezing non-targeted parameters at their initial values $\Theta_0$:
\begin{itemize}
    \item Input-Hidden: Only layer 1 weight matrix $W^{(1)}$ and bias $b^{(1)}$ evolve over stream.
    \item Hidden-Hidden: Only intermediate layers change while input and output layers are fixed.
    \item Hidden-Output: Only output layer weight matrix $W^{(3)}$ and bias $b^{(3)}$ change over the stream.
\end{itemize}

Fig.~\ref{fig:appendix_experiments}a demonstrates the effect of changing the label space by shifting Hidden-Output weights versus shifting the input or hidden representational space of the teacher.
Consistent with readout-similarity effects in \cite{lee2021continual}, we see that shifts in the label space cause the learner to forget more than shifts in the input or hidden representational spaces of the teacher.

\subsection{Stream scope}

To test how the scope or the support of the task stream influences student forgetting, we expand the intrinsic dimensionality of the weight trajectory manifold $\mathbf{w}(t)$ from 1D to 3D.
Added dimensions are generated by random Brownian motion multiplied by a linear decay envelope $(1 - t)$ to make sure that the trajectory ends at the exact same parameter target at $t = 1$ as it started with at $t=0$.
We made this choice to visualize how similarity between tasks does not necessarily depend on temporal adjacency in the stream; however other choices could have been made.

Fig.~\ref{fig:appendix_experiments}b demonstrates the effect on the learner of changing the teacher weight trajectory's intrinsic dimensionality, and thus the stream scope.
The only clear qualitative trend that shines through is that the higher dimensional trajectories seem harder to retain than the simple 1D trajectory projected into the teacher's weight space. We acknowledge that increasing the intrinsic dimensionality of the trajectory does not change scope in isolation, but also lengthens the path the teacher traverses, alters pairwise distances between tasks and changes the local rate of drift. The observed effect should therefore be read as the effect of this joint change rather than of scope alone. 

\subsection{Student architecture and evaluation}

The student network is an MLP with layers $[10, 128, 128, 2]$ trained with Adam ($\eta = 0.01$, batch size 32) for 10 epochs per task, evaluated across 5 independent random initializations. 
We report only the final stream loss ($L_{N-1,n}$), which is the MSE test loss on task $n$ evaluated after the student model has completed training on the complete task stream.

\end{document}